\documentclass{article}[12pt]
\usepackage[utf8]{inputenc}
\usepackage[export]{adjustbox}
\usepackage{amsmath,amssymb,graphicx}
\usepackage{multirow}
\usepackage{citesort}

\newcommand{\bcs}{\begin{cases}}
\newcommand{\bi}{\begin{itemize}}
\newcommand{\bc}{\begin{center}}
\newcommand{\be}{\begin{enumerate}}
\newcommand{\beq}{\begin{equation}}
\newcommand{\bbm}{\begin{bmatrix}}
\newcommand{\ec}{\end{center}}
\newcommand{\ei}{\end{itemize}}
\newcommand{\ee}{\end{enumerate}}
\newcommand{\ecs}{\end{cases}}
\newcommand{\eeq}{\end{equation}}
\newcommand{\ebm}{\end{bmatrix}}

\newcommand{\bal}{\begin{aligned}}
\newcommand{\eal}{\end{aligned}}
\newcommand{\rt}{\rightarrow}

\newcommand{\bx}{\mathbf{x}}

\newcommand{\bu}{\mathbf{u}}

\newcommand{\nid}{\noindent}

\def\bv{{\bf v}}
\def\bb{{\bf b}}
\def\bc{{\boldsymbol{\xi}}}

\def\bw{{\bf w}}
\def\nid{\vspace{1ex}\noindent}
\def\nidd{\vspace{2ex}\noindent}

\usepackage[letterpaper,top=2cm,bottom=2cm,left=3cm,right=3cm,marginparwidth=1.75cm]{geometry}

\title{Characteristics-Informed Neural Networks for\\ Forward and Inverse Hyperbolic Problems}

\author{Ulisses Braga-Neto \\
        \small Department of Electrical and Electronic Engineering\\
        \small Texas A\&M University\\
        \small E-mail: \tt{ulisses@tamu.edu} 
}
\date{}

\begin{document}
\maketitle
\begin{abstract} 
  \noindent We propose characteristics-informed neural networks (CINN), a simple and efficient machine learning approach for solving forward and inverse problems involving hyperbolic PDEs. Like physics-informed neural networks (PINN), CINN is a meshless machine learning solver with universal approximation capabilities. Unlike PINN, which enforces a PDE softly via a multi-part loss function, CINN encodes the characteristics of the PDE in a general-purpose deep neural network by adding a characteristic layer. This neural network is trained with the usual MSE data-fitting regression loss and does not require residual losses on collocation points. This leads to faster training and can avoid well-known pathologies of gradient descent optimization of multi-part PINN loss functions. This paper focuses on linear transport phenomena, in which case it is shown that, if the characteristic ODEs can be solved exactly, then the output of a CINN is an exact solution of the PDE, even at initialization, preventing the occurrence of non-physical solutions. In addition, a CINN can also be trained with soft penalty constraints that enforce, for example, periodic or Neumman boundary conditions, without losing the property that the output satisfies the PDE automatically. We also propose an architecture that extends the CINN approach to linear hyperbolic systems of PDEs. All CINN architectures proposed here can be trained end-to-end from sample data using standard deep learning software.  Experiments with the simple advection equation, a stiff periodic advection equation, and an acoustics problem where data from one field is used to predict the other, unseen field, indicate that CINN is able to improve on the accuracy of the baseline PINN, in some cases by a considerable margin, while also being significantly faster to train and avoiding non-physical solutions. An extension to nonlinear PDEs is also briefly discussed.
\end{abstract}

\section{Introduction}

Physics-informed neural networks (PINNs) are revolutionizing the application of machine learning methods in engineering and science, by introducing physics laws to complement data in training predictive models~\cite{karniadakis2021physics}. 
A standard neural network is agnostic to the physics of the problem, and therefore requires a large amount of data distributed over the entire domain in order to make reliable predictions. On the other hand, the physics-informed  approach includes the physics in the training of the neural network and therefore not only requires less data, but also has the ability to extrapolate to regions of the domain where there are no data. In addition, PINNs do not require elaborate meshes, make predictions at any given test points in an irregular spatial-temporal domain (rather than producing a solution over a pre-specified grid), are ideally suited for inverse problems, and can be trained using standard state-of-the-art deep learning software~\cite{raissi2019physics}.

Despite its many advantages, the PINN approach is plagued by training issues, created in large measure by a multi-part loss function that attempts to constrain the neural network to simultaneously fit the data and satisfy the PDE, which is a hard multi-objective optimization problem. Diverse attempts have been made to mitigate this issue, including weighting the loss function~\cite{wang2022and,mcclenny2020self,van2021optimally} and decomposing the solution domain~\cite{jagtap2020extended,shukla2021parallel}. In addition to the training issue, solutions produced by PINNs are often non-physical, due to the soft constraint implemented by the loss function.
Ideally, all these issues would be resolved by hard-encoding the physics, represented by a PDE, directly into the architecture of the neural network. This would both avoid the occurrence of non-physical solutions and allow the neural network to be trained with a standard data-fitting loss function. This network would maintain all the advantages of PINNs, such as working with small amounts of data, requiring no meshes, and making predictions at any point in a possibly irregular domain.

Here, such a neural network architecture is proposed, in the important case of hyperbolic PDEs. Such PDEs describe wave-like phenomena, which occur in nearly all fields of science and engineering~\cite{leveque2002finite}. We note that PINNs have been extensively studied in hyperbolic problems; e.g. see ~\cite{abreu2021study,jagtap2020conservative,patel2022thermodynamically,fraces_physics_2021,fraces_physics_2020,fuks_limitations_2020,coutinho2022physics}. First-order hyperbolic PDEs can be solved with the method of characteristics (higher-order PDEs can often be rewritten as systems of first-order PDEs, which can likewise be solved). 
The proposed {\em characteristics-informed neural network} (CINN) is a simple, general deep neural network architecture that takes advantage of characteristics to solve forward and inverse hyperbolic problems. In important cases, the characteristic curves can be determined exactly by solving the characteristic ODEs analytically; in those cases, we show that
the output of a CINN is an exact solution of
the PDE, even at initialization, preventing the occurrence of non-physical
outputs. Otherwise, approximate numerical methods can be used to solve the characteristic ODEs, leading to approximate solutions of the PDE, but the CINN is still trained quickly and efficiently using only
a data-fitting loss function. In this paper, we first demonstrate how to use CINNs to forward and inverse problems involving liner transport PDEs, in particular, the simple advection equation. Next, we show that CINNs can be also trained with soft penalty constraints as well, without losing the property that the output satisfies the PDE automatically; here, we consider enforcing periodic boundary conditions for the advection problem. Finally, we propose an architecture that extends CINN to linear systems of PDEs. The preprint also outlines a future extension to nonlinear hyperbolic PDEs.

In forward and inverse problem experiments with the simple advection equation, the CINN proved to be about as accurate as a comparable baseline PINN, while being nearly twice as fast (and avoiding non-physical solutions). More complex experiments with the periodic advection equation and a ``hidden'' acoustics problem show that, as the problems become more difficult, CINN displays significantly more accurate results than a comparable baseline PINN, while still being faster, and avoiding non-physical solutions. The periodic advection experiments allows us to study the effect of stiffness on training, by varying the speed of the periodic advection; as the speed increases, the problem becomes stiffer. We observed that stiffness slows down converge to the solution, but that CINN is less affected by this than the baseline PINN. Indeed, in some cases when the PINN did not converge, the CINN was able to obtain an accurate solution. On the other hand, the purpose of the hidden acoustics experiment is to use the classical equations of linear acoustics and scattered data on one of the fields (in this case, the velocity), to predict the other, unseen field (in this case, the pressure). In this more complex experiment, we observed that the CINN was significantly more accurate than the baseline PINN; the latter required five times as many training iterations to match the accuracy of the velocity prediction, while still not matching the accuracy of the unseen pressure prediction.

This paper is organized as follows. Section 2 is a survey of related work, with an emphasis on methods that include hard constraints in neural network architectures to satisfy physical constraints. Section 3 contains definitions and background material on neural networks, PINNs, and hyperbolic PDEs, including the transport equation. Section 4 is the main section in the paper, where CINN architectures for scalar and systems of PDEs are introduced, and experimental results using several PDEs are presented. Section 5 discusses briefly an extension of the CINN approach to nonlinear hyperbolic PDEs. Finally, Section 6 provides concluding remarks.

\section{Related work}

Many methods have been proposed to include physical knowledge directly into the architecture of neural networks. The most well-known is likely the method used in \cite{lagaris1998artificial} to modify the {\em output} of a general neural network to make initial or boundary conditions automatically satisfied. For example, if $u_\theta(x,t)$ is the output of the neural network, and it is known that the solution of the problem satisfies an initial condition $u(x,0) = g(x)$, then the conditioned output
\beq
  \tilde{u}_\theta(x,t) \,=\, u_\theta(x,t)t + g(x)e^{-t}
\label{eq-lagaris}
\eeq
satisfies $\tilde{u}_\theta(x,0) = g(x)$ automatically (a similar approach can be applied to  spatial boundary conditions). Alternatively, the initial condition $g(x)$ might be unknown and itself approximated by a separate neural network, as in \cite{sheng2021pfnn}. Using \eqref{eq-lagaris} works well in many cases, though it tends to make the neural network $u_\theta(x,t)$  more difficult to train. More importantly, it does not allow a PDE to be satisfied automatically, but only initial or boundary conditions. To include the PDE, the neural network still has to be trained using a collocation grid and a soft penalty term in the loss function.

In \cite{mattheakis2019physical}, a similar output-conditioning approach is employed to enforce odd/even symmetry. The idea is based on the observation that $f(x) + f(-x)$ and $f(x) - f(-x)$ are even and odd functions, respectively, regardless of the function $f(x)$. For example, to enforce odd spatial symmetry, the conditioned output would be
\beq
  \tilde{u}_\theta(x,t) \,=\, u_\theta(x,t) \,+\, b(x)\,,
\eeq
where the ``bias term'' is $b(x) = -1/2[u_\theta(x,t)+u_\theta(-x,t)]$.
  
In \cite{dong2021method}, a simple approach is proposed that modifies the {\em input} of the neural network to ensure that periodic boundary conditions are automatically satisfied. This method is based on the observation that if $g(x)$ is periodic of period $L$, then the composition $f(g(x))$ is also periodic of period $L$, no matter that $f$ is. The idea is then to let 
\beq
  \tilde{u}_\theta(x,t) \,=\, u_\theta(g(x),t)\,,
\eeq
where $u_\theta(x,t)$ is s general neural network, and, for example, $g(x) = A\cos(2\pi x/L)$. This is reminiscent of the CINN approach proposed here, in that it modifies the input of the neural network. 

Positivity and monotonicity constraints can be implemented by constraining the weights and activation functions of the neural network~\cite{sill1997monotonic}. An interesting approach to monotone neural networks applies a positivity constraint and then integrates the output of the neural network to obtain an increasing function~\cite{wehenkel2019unconstrained}. Such approaches have been applied to physics-informed neural networks~\cite{zhang2022monotonicity}.

Most PINN research has been based on standard fully-connected neural network architectures. Modified architectures have been proposed to accelerate training or avoid complete lack of convergence. These include the use of skip connections in \cite{zhang2022implicit}, the addition of radial-basis function layers in \cite{xiao2022hybrid}, and the use of transformer encoders in \cite{wang2021understanding}. These modified architectures are however not informed directly by the PDE.

Attempts have been made to encode traditional finite-difference methods into neural network architectures for solving PDEs. This include the finite-difference convolutional neural network layers in ~\cite{shi2020finite,kim2022learning} and the discrete PINN in \cite{raissi2019physics}. These methods however inherit the problems that traditional discretization-based solvers face. In~\cite{darbon2021some}, a class of Hamilton-Jacobi PDEs are solved by implementing the Lax-Oleinik formula as a neural network. However, this is not a general-purpose neural network architecture, nor can it be trained using data.

An interesting deep learning approach to solve linear or nonlinear transport PDEs is proposed in \cite{zhang2022implicit}, where instead of enforcing the PDE as in PINN, the implicit form of the solution is enforced instead. However, no attempt is made to include the physics into the neural network architecture, but rather a soft penalty constraint is used. The method also requires knowledge of the solution at the initial boundary $t=0$ (incidentally, this could be replaced by knowledge of the solution along any non-characteristic curve). A more closely related approach to CINN is found in \cite{mojgani2022lagrangian}, where the authors do propose a specialized neural network architecture for nonlinear advection-diffusion problems, which computes the characteristic curves and the value of the solution along those curves. However, crucially, the physics is still enforced by soft penalty constraints. In addition, an additional numerical interpolation step is needed to produce a solution in the original variables, so the neural network cannot be trained end-to-end with arbitrary data.

All previous approaches either enforce general properties of the neural network output using hard constraints (but not the PDEs themselves), or use soft penalty constraints. By contrast, CINN is a simple and general approach that embeds the physics of hyperbolic problems into the neural network and does not rely on soft penalty constraints. The modification made to the neural network architecture is a simple addition of a characteristic layer, which still allows it to be trained end-to-end using standard, popular deep learning software, such as the Keras API of Tensorflow 2.x~\cite{chollet2015keras,tensorflow2015-whitepaper}, even in the more complex case of PDE systems. 

\pagebreak
\section{Background}

In this section, we set the notation and introduce the necessary concepts used in the paper.

\subsection{Neural Networks}

Neural networks provide a data-driven approach to estimating an unknown function $u:R^d \rt R$ (a scalar unknown is assumed here for simplicity; the extension to the vectorial case is straightforward). For this purpose, a data set $S_n = \{(\bx_1,y_1),\ldots,(\bx_n,y_n)\}$ is used, where $\bx_i \in R^d$ and $y_i = u(\bx_i) + \epsilon_i$, $i=1,\ldots,n$. The sensor noise random variables $\epsilon_i$ are usually modeled as independent zero-mean Gaussian, however other choices are possible. In addition, some observations could be noiseless ($\epsilon_i=0$), e.g., data coming from known boundary conditions. There could be also observations on the derivatives of $u$, but this case is not considered here. 

A fully-connected neural network $u_\theta(\bx)$ with $L$ layers is a function $u_\theta:R^d \rightarrow R$ defined by
\begin{equation}
    \label{eq:fully-connectedNN}
    u_\theta(\bx) \,=\, W^{L-1}\sigma 
                ( \cdots \sigma (W^{0}\bx + \bb^{0}) + \cdots) + \bb^{L-1},
\end{equation}
where $\sigma$ is an entry-wise differentiable activation function, $W^{l}$ and $\bb^{l}$ are respectively the weight matrix and the bias vector for layer $l$, for $l=0,\ldots,L-1$, and $\theta =  (W^{0}, \bb^0,\ldots,
    W^{L-1}, \bb^{L-1})$ is the vector of parameters. Neural network training consists in tuning the parameter vector $\theta$ to minimize the mean-square loss function:
\begin{equation}
    \mathcal{L}(\theta) \,=\, \frac{1}{n}\sum_{i=1}^{n} |y_i - u_\theta(\bx_i)|^2.
\label{eq-L}
\end{equation}
The method of choice in training {\em deep} neural networks ($L \gg 1$) is gradient descent from a randomly initialized $\theta$. Sophisticated methods using adaptive learning rates and momentum are typically used to perform gradient descent efficiently~\cite{geron2019hands}.

The quality of approximation provided by a neural network is usually assessed by the $L_p$ error:
\begin{equation}
    L_p \ \text{error} \,=\, \frac{\left(\sum_{i=1}^{m} |u_\theta(\bx_i) - u(\bx_i)|^p\right)^{1/p}}{\left(\sum_{i=1}^{m} |u(\bx_i)|^p\right)^{1/p}}\,,
\end{equation}
where $p=1,2$ and $\bx_1,\ldots,\bx_m$ comprise a sufficiently dense set of test points.

\subsection{Physics-Informed Neural Networks}

Now, suppose that the unknown $u$ satisfies a PDE 
\beq
\mathcal{N}[u](\bx) \,=\, 0\,, \ \ \bx \in \Omega\,,
\eeq
where $\mathcal{N}$ is a differential operator. In the physics-informed neural network (PINN) approach \cite{raissi2019physics}, the neural network is encouraged to satisfy the PDE by introducing a residue loss function
\begin{equation}
    \mathcal{L}_r(\theta) \,=\, \frac{1}{n_r}\sum_{i=1}^{n_r} |\,\mathcal{N}[u_\theta](\bx_r^i)|^2,
\label{eq-Lr}
\end{equation}
where the ``collocation'' points $\bx_r^1,\ldots,\bx_r^{n_r}$ are randomly distributed in the domain $\Omega$. Application of the operator $\mathcal{N}$ to the neural network function $u_\theta$ can be performed efficiently using automatic differentiation methods~\cite{baydin2018automatic}. The loss minimized in PINN training is the sum of the data-fitting loss and the PDE residual loss  
\begin{equation}
    \mathcal{L}_{\rm PINN}(\theta) \,=\, \mathcal{L}(\theta) + \mathcal{L}_r(\theta)\,.
\end{equation}
The loss terms could be weighted to give different importance to the data-fitting and residue components: $\mathcal{L}_{\rm PINN}(\theta) \,=\, \mathcal{L}(\theta) + \lambda \mathcal{L}_r(\theta)$, where the weight $\lambda$ can be fixed or adjusted during training~\cite{wang2021understanding}. Alternatively, self-adaptive weights could be assigned to each data or collocation point~\cite{mcclenny2020self}. 

\subsection{Hyperbolic PDEs}

Hyperbolic PDEs may be roughly defined as ``equations supporting wave-like solutions''~\cite[Sec.~7.3]{evans2010partial}. Hyperbolic PDEs often arise from a multidimensional conservation law:
\beq
   \frac{\partial u(\bx,t)}{\partial t} \:+\: {\rm div}\, {\bf F}(\bx,t,u) \,=\, 0\,,
\label{eq:conserv}
\eeq
which states that the instant time variation of the unknown $u$ must exactly match the instant spatial divergence of the {\em flux} ${\bf F}$, i.e., $u$ is conserved. 

Depending on how the flux is defined, a wide array of physical phenomena can be modeled. In this paper, we mostly deal with the {\em continuity equation}, in which case $u$ stands for the spatial density of a conserved quantity (such as mass, momentum, or energy) and ${\bf F} = u\,\bv$, where $\bv(\bx,t,u) = (v_1(\bx,t,u),\ldots,v_d(\bx,t,u))$ is a velocity vector field. By using the divergence product rule ${\rm div}(u\bv) = \bv \cdot \nabla u + u\,{\rm div}\, \bv$, we can write \eqref{eq:conserv} as
\beq
   \frac{\partial u(\bx,t)}{\partial t} \:+\:  \bv(\bx,t,u)\cdot \nabla u(\bx,t) \,=\, -u(\bx,t)\,{\rm div}\,\bv(\bx,t,u)\,,
\label{eq:tr1}
\eeq
where the gradient is along the spatial coordinates only. If ${\rm div}\,v(\bx,t,u) = 0$, i.e., the flow is {\em divergence-free} (also known as an ``incompressible'' flow), then the source term disappears, and one obtains the {\em transport equation}:
\beq
   \frac{\partial u(\bx,t)}{\partial t} \:+\:  \bv(\bx,t,u)\cdot \nabla u(\bx,t) \,=\, 0\,,
\label{eq:tr2}
\eeq
which governs the concentration $u(\bx,t)$ of a substance moving in a fluid with velocity field $\bv(\bx,t,u)$. If the flow is not divergence-free, then the transport equation still has physical meaning, but the quantity $u(\bx,t)$ is not conserved. 

Notice that, in a more complex situation, the fluid velocity field itself could be determined by a PDE, but here we assume it is either given (in the forward problem), or it is estimated from data (in the inverse problem). Also notice that if the velocity field is only a function of space and time, but not of $u$ itself, then the PDEs \eqref{eq:tr1} and \eqref{eq:tr2} are linear. This would be the case, for example, where the concentration $u$ is small enough as not to disturb the fluid velocity field. Otherwise, the previous PDEs are nonlinear (quasi-linear), in which case it is well-known that the solution may not exist for all times in a classical sense, due to the occurrence of ``shocks''~\cite{leveque2002finite}. Finally, the previous PDEs apply to a scalar unknown, but they can be extended to the vector case, i.e., systems of PDEs. 


\section{Characteristics-Informed Neural Networks}

We present below a methodology to build neural network architectures that will automatically satisfy a hyperbolic PDE, rendering the residual loss function \eqref{eq-Lr} obsolete in this case. Training of such neural networks uses only the data-fitting loss \eqref{eq-L}, as an ordinary neural network would (though extra loss terms could be used, as shown in Section~\ref{Sec-periodic}). The PDE is guaranteed to be satisfied over the entire domain, not only a finite set of collocation points, so that non-physical approximations cannot occur. 

In this manuscript, we focus on the multidimensional linear transport equation, both in the scalar and vectorial (PDE system) cases. An extensions to  nonlinear PDEs is discussed in Section~\ref{Sec-nonlin}. More general cases, including equations with sources, such as the compressible flow in \eqref{eq:tr1}, will be addressed in future work.

\subsection{Linear Scalar PDE}
\label{Sec-linscalar}

Consider the linear, multi-dimensional, variable-coefficient version of the transport equation in \eqref{eq:tr2}: 
\beq
   \frac{\partial u(\bx,t)}{\partial t} \:+\: \bv(\bx,t)\cdot \nabla u(\bx,t) \,=\, 0\,,
\label{eq:ls}
\eeq
where the velocity vector field $\bv(\bx,t)$ is assumed to be sufficiently smooth. For example, in the case $d=1$, $u(x,t)$ could represent the time-varying concentration in grams/m$^3$ of a tracer at point~$x$ along a pipe. Equation~\ref{eq:ls} is known as the ``color equation'' in this context~\cite{leveque2002finite}, since the volumetric concentration $u$ determines the color of the fluid. The tracer moves along with the fluid with varying velocity $v(x,t)$ along the pipe. In the linear case, the tracer concentration $u(x,t)$ is assumed to be small enough so as not to disturb $v(x,t)$, making the latter a function of only $x$ and $t$, but not $u$. In the forward problem, the velocity $v(x,t)$ is assumed to be known, whereas in an inverse problem, it could be unknown and be sought through experimental measurements of the tracer concentration at various points in the pipe.

A key property of equation \eqref{eq:ls} is that it can be solved along characteristic curves in the spatial-temporal domain. The characteristic curve $\bx(t)$; $t \geq 0$ satisfies the ODE:
\beq
  \frac{d \bx(t)}{dt} \,=\, \bv(\bx(t),t)\,.
\label{eq:charac}
\eeq
This is really a system of $d$ ODEs:
\beq
  \frac{d x_i(t)}{dt} \,=\, v_i(\bx(t),t)\,, \quad i=1,\ldots,d\,.
\label{eq:charac2}
\eeq
Clearly, along a characteristic curve $\bx(t)$, one has
\beq
\bal
  \frac{du(\bx(t),t)}{dt} & \,=\,  \frac{\partial u(\bx(t),t)}{\partial t} + \frac{d \bx(t)}{dt} \cdot \nabla u(\bx(t),t) \\
  & \,=\, \frac{\partial u(\bx(t),t)}{\partial t} + \bv(\bx(t),t) \cdot \nabla u(\bx(t),t)\\
  & \,=\, 0
\eal
\eeq
by virtue of \eqref{eq:ls} and \eqref{eq:charac}. Therefore, $u(\bx(t),t)$ is constant along the characteristic $\bx(t)$.

Let a solution of \eqref{eq:charac} be written in the implicit form
\beq
  \phi(\bx,t) = \bc\,,
\label{eq:chtf}
\eeq  
where $\bc \in R^d$ is the constant of integration of the ODE system. The {\em characteristic transform} $\phi(\bx,t)$ maps the original problem variables $\bx,t$ into the characteristic vector $\bc$. As $\bc$ ranges over $R^d$, all the characteristic curves are described by \eqref{eq:chtf}. Furthermore, a solution of \eqref{eq:ls} must be constant over each characteristic curve. This suggests the following fundamental result.

\vspace{1.5ex}
\noindent
{\bf Theorem.} A solution of \eqref{eq:ls} is of the form
\beq
  u(\bx,t) \,=\, f(\phi(\bx,t))\,,
\eeq
where $f:R^d\rt R$ is an arbitrary differentiable function.

\vspace{1.5ex}
\noindent
{\em Proof.} Since the velocity field is smooth everywhere, the Picard-Lindel\"of theorem \cite{tenenbaum1985ordinary} guarantees that a unique solution of \eqref{eq:charac} passes through any given point $(\bx_0,t_0)$ in the spatial-temporal domain. The characteristic curves therefore partition the domain as $\bc$ ranges over $R^d$. Since a solution of \eqref{eq:ls} must be constant over each characteristic curve, the result follows.$\:\Box$  
\vspace{1.5ex}

\begin{figure}[t!]
\centering
\includegraphics[scale=0.7]{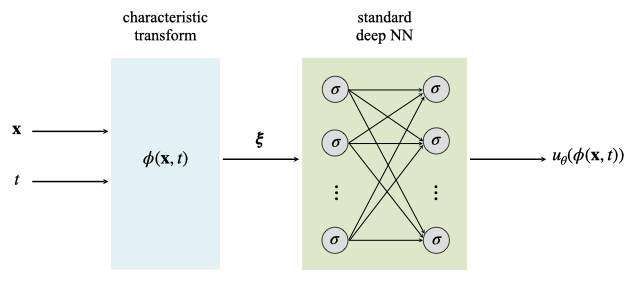}
\caption{CINN architecture for a hyperbolic PDE describing a linear incompressible flow with varying velocity field. The characteristic transform  $\phi(\bx,t)$ maps the space and time input variables $\bx,t$ into the characteristic vector $\bc$. The output of the neural network automatically  satisfies the PDE.}
\label{Fig-CINN_linear}
\end{figure}

The basic idea behind CINNs is to take advantage of this key result by expressing the unknown function $f$ as a standard neural network $u_\theta$, and applying it to the transformed input $\bc = \phi(\bx,t)$. Theorem~1 guarantees that the output $u_\theta(\bc) = u_\theta(\phi(\bx,t))$ of the CINN must satisfy the PDE \eqref{eq:ls} automatically. After the transformation layer, the neural network is trained to compute the appropriate value of the solution along the characteristics in order to satisfy any initial/boundary condition data or experimental measurements. Only a standard data-fitting loss term is used, since the PDE is satisfied automatically by the CINN, and no soft penalty terms are needed. See Figure~\ref{Fig-CINN_linear} for an illustration.

The PDE is satisfied exactly provided that the transformation $\bc = \phi(\bx,t)$ can be determined analytically. If that is not the case, then it must itself be approximated, either by using a PINN, a physics-informed Gaussian process, or other continuous numerical method. In that case, the PDE is satisfied approximately by the CINN, where the degree of accuracy is a function of the ODE solver accuracy. 

\subsubsection{Advection with Constant Velocity}

In the simple case $\bv(x,t) \equiv \bv$, \eqref{eq:ls} reduces to the ``advection equation'' (with constant velocity),
\beq
   \frac{\partial u(\bx,t)}{\partial t} \:+\: \bv\cdot \nabla u(\bx,t) \,=\, 0\,,
\label{eq:lss}
\eeq
which models uniform transport along the direction pointed to by $\bv$ with speed $||\bv||$.
This velocity field is clearly divergence-free, so that the advection equation \eqref{eq:lss} is a conservation law, and $u$ is a conserved quantity (see Section~3.3). As an aside, this is the only possible divergence-free velocity field in the case $d=1$; any other case leads to a nonconservative transport equation.

\begin{figure}[t!]
\centering
\includegraphics[scale=0.68]{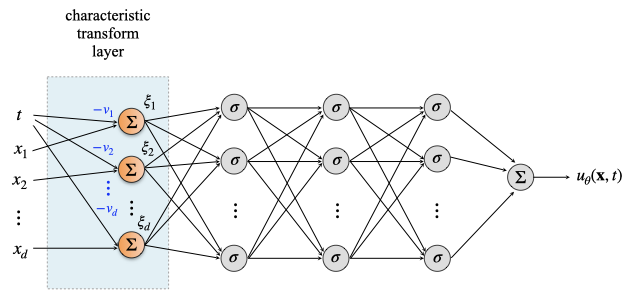}
\caption{CINN architecture for the advection equation. The velocity field components are weights in the first layer. These weights are fixed in the forward problem, but trainable in the inverse problem.}
\label{fig:cinn_advection}
\end{figure}

The characteristic equation \eqref{eq:charac} in this case reads
\beq
  \frac{d \bx(t)}{dt} \,=\, \bv\,,
\label{eq:characs}
\eeq
the solution of which is clearly
\beq
  \bx(t) \,=\, \bv t + \bc
\eeq
where $\bc \in R^d$ is the constant of integration. The characteristic transform in this case takes the following simple form:
\beq
  \phi(\bx,t) \,=\, \bx - \bv t \,=\, \bc\,,
\eeq
and the general solution of the transport PDE is, according to Theorem~1, of the form
\beq
  u(\bx,t) \,=\, f(\bx-\bv t)\,.
\label{eq:adv_sol}
\eeq  
The solution therefore takes constant values on the parallel characteristic lines $\bx - \bv t = \bc$. 
The characteristics-informed neural network in this case is depicted in Figure~\ref{fig:cinn_advection}. The characteristic transform layer is a regular neural network layer, except that the weights multiplying the spacial components are always equal to $1$. The velocity components are the negatives of the weights multiplying the time input. In a forward problem, these weights are known and non-trainable, while in the inverse problem, they are trainable, to allow the velocity to be determined from the data.

\nidd
\underline{Forward Problem}
  
\nid
Let us consider the advection equation in the case $d=1$, with a ``Riemann'' initial condition:
\begin{align}
  & \frac{\partial u(x,t)}{\partial t} \:+\: v\, \frac{\partial u(x,t)}{\partial x} \,=\, 0\,, \quad (x,t) \in [0,L] \times [0,T]\,,\\
  & u(x,0) = g(x) = \begin{cases} u_l\,, & x < L/2\,,\\ u_r\,, & x > L/2\,,
  \end{cases}\\
  & u(0,t) = u_l\,, \:\: u(L,t) = u_r\,.\label{eq:initial}
\end{align}
Letting $t=0$ in \eqref{eq:adv_sol}, we conclude that $f(x) = g(x)$, and so the solution of the PDE is
\begin{equation}
     u(x,t) \,=\, g(x-vt)\,, \ \ x \in R \,, \ t > 0\,, 
\end{equation} 
i.e., the initial step profile is simply translated to the right at constant speed $v$. In our experiment, we set $L=2$, $v=1$, and $T=0.8$. The final time $T$ is selected so that $vT < L/2$, so that the initial discontinuity does not move past the end of the interval. This assumption could be relaxed, at the expense of introducing a more complex boundary condition.

\begin{figure}[t!]
\centering
\includegraphics[scale=0.25]{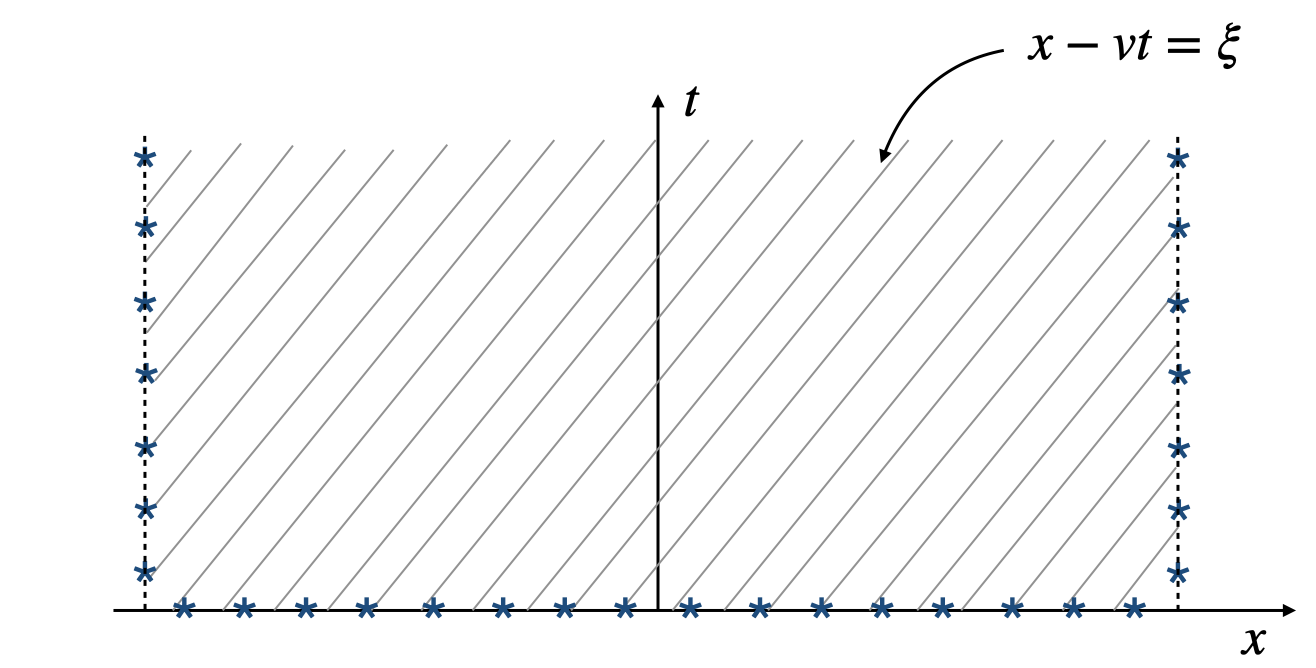}
\caption{The characteristic curves for the one-dimensional advection equation, along with sample data from initial and boundary conditions used in the forward problem. A solution of the PDE must have a constant value along each characteristic line. The CINN learns these values from the sample data, but cannot change the characteristic structure itself. Hence, the output of the CINN, even prior to training, is a valid solution of the PDE.} 
\label{fig:advection_ch_curve}
\end{figure}

In the forward problem, the initial and boundary conditions are sampled to provide the data to train the CINN. See Figure~\ref{fig:advection_ch_curve} for an illustration of the characteristic structure and sample data from initial and boundary conditions. From the previous discussion, we know that a solution of the PDE must have a constant value along each characteristic line. The CINN learns these values from the sample data, but cannot change the characteristic structure itself. Hence, the output of the CINN, even prior to training, is a valid solution of the PDE.  

\begin{figure}[t!]
\centering
\begin{tabular}{c}
CINN after 100 iterations\\
\includegraphics[scale=0.3]{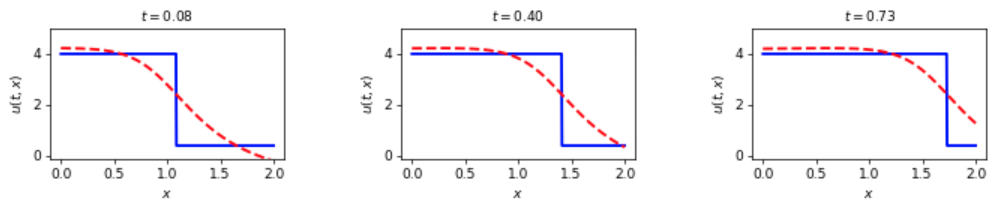}\\[2ex]
CINN after 200 iterations\\
\includegraphics[scale=0.3]{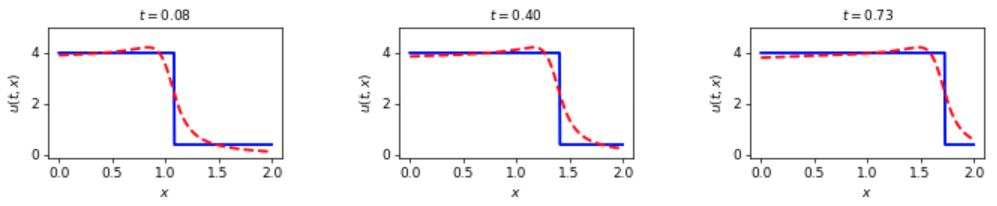}\\[2ex]
CINN after 1000 iterations\\
\includegraphics[scale=0.3]{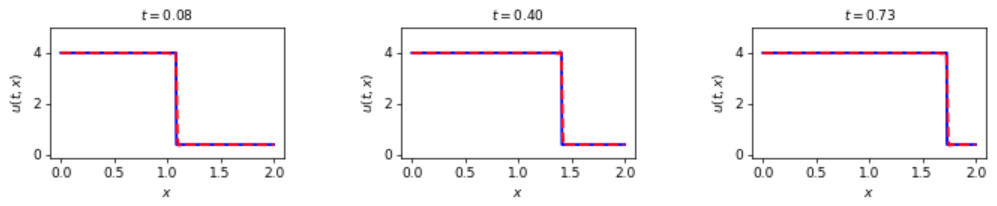}
\end{tabular}
\label{fig:advection_CINN_results}
\caption{Typical CINN forward prediction results for the advection PDE. The CINN output is a solution of the advection PDE at any point during training, and conservation of mass is never violated.}
\end{figure}

In our experiment, we employ 50 sample data points along the initial boundary $t=0$, and only 5 data points in each lateral boundary $x=0$ and $x = L$. The points are uniformly spaced along each boundary. There are no collocation points or residual loss. The CINN consists of the initial characteristic layer, followed by 8 layers of 20 neurons each, and the output layer, with hyperbolic tangent nonlinearity, and Glorot initialization for the trainable weights (the weights in the characteristic layer are non-trainable, and fixed at $-v = -1$ and $1$). The network is trained with the standard MSE data-fitting loss function, using ADAM gradient descent with the usual parameters, and a constant learning rate of $0.001$. Typical results can be seen in Figure~\ref{fig:advection_CINN_results}. We can observe that convergence after 1000 training iterations is very good. Critically, we observe that at any time during training, the CINN output is a solution of the advection PDE, as it consists of the same profile translating to the right at the correct velocity. Hence, conservation of mass is never violated.

\begin{figure}[t!]
\centering
\begin{tabular}{cc}
NN after 100 iterations & \hspace{1em}PINN after 100 iterations \\
\includegraphics[scale=0.2]{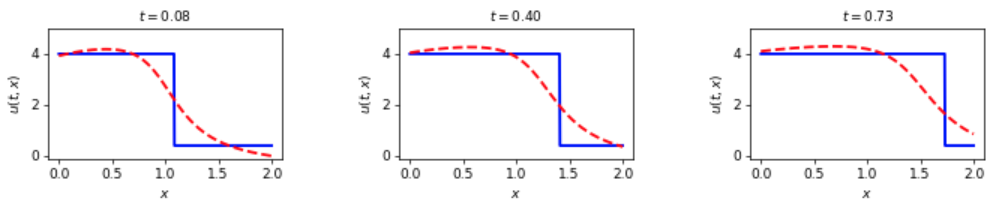} & \hspace{1em}
\includegraphics[scale=0.2]{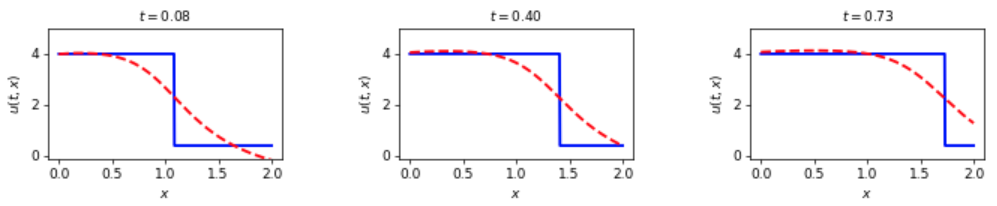}\\[2ex]
NN after 200 iterations & PINN after 200 iterations \\
\includegraphics[scale=0.2]{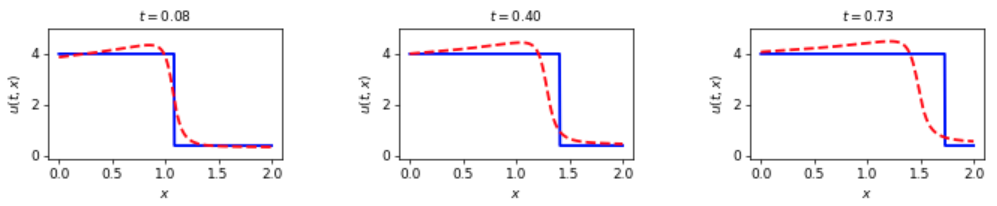} & \hspace{1em}
\includegraphics[scale=0.2]{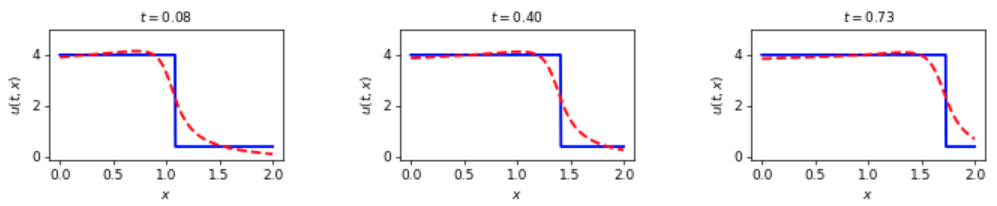}\\[2ex]
NN after 1000 iterations & PINN after 1000 iterations \\
\includegraphics[scale=0.2]{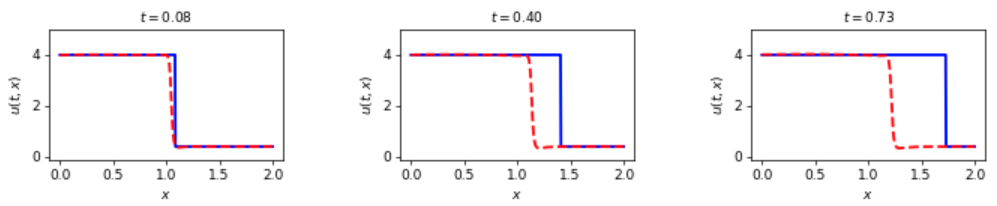} & \hspace{1em}
\includegraphics[scale=0.2]{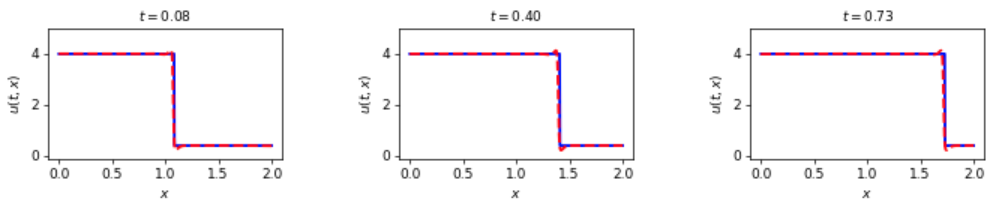}
\end{tabular}
\label{fig:advection_NN_PINN_results}
\caption{Typical NN and PINN forward prediction results for the advection PDE. The NN cannot extrapolate the solution into later times, while the PINN cannot guarantee that conservation of mass is never violated.}
\end{figure}

For comparison, the results obtained by a standard neural network and a standard PINN, using the same hyperparameters and training data as in the CINN case, including the same neural network architecture (minus the characteristic layer), are displayed in Figure~\ref{fig:advection_NN_PINN_results}. In the PINN case, 300 collocation points were sampled in the spatial-temporal domain using Latin hypercube sampling. We can observe that the standard NN can fit the data to some extent at early times, where most of the data are, but it is unable to track the correct solution at later times, where there is only lateral boundary data. The PINN naturally does a much better job than the NN, but we can observe that the PINN solution is non-physical: it is not a translation to the right of the same profile, that is, mass conservation is violated; this is naturally the case of the NN as well (in both the NN and PINN cases, the violation can be observed more easily at 100 iterations, on the left of the plots).

For a quantitative comparison, we ran each experiment independently 10 times using each time different random initialization of the neural network weights, and recorded the average $L_1$ error, $L_2$ error (as defined previously), and execution time. The results are displayed in Table~\ref{tab:advection_forward}. We can see that CINN and PINN have comparable accuracies in this simple problem, and both are much more accurate than the simple NN. However, the CINN is almost twice a fast the PINN, despite the fact that the neural network has an extra layer. 

\begin{table}[h!]
\centering
\begin{tabular}{c|ccc}
\hline
Solver & $L_1$ error & $L_2$ error & running time (sec)\\
\hline
NN & $0.1251 \pm 0.0649$ & $0.3120 \pm 0.1066$ & $5.9388 \pm 0.9181$\\[1ex]
PINN & $0.0118 \pm 0.0066$ & $0.0619 \pm 0.0275$ & $10.2154 \pm 1.1794$\\[1ex]
CINN & $0.0160 \pm 0.0094$ & $0.0550 \pm 0.0265$ & $5.7178 \pm 0.1349$\\
\hline
\end{tabular}
\label{tab:advection_forward}
\caption{Advection forward prediction results over 10 independent repetitions.}
\end{table}

\nidd
\underline{Inverse Problem}

\nid
We compare the PINN and CINN approaches to the inverse problem by assuming that the constant advection velocity is to be estimated from noisy experimental data. In the CINN approach, the velocity appears in the neural network architecture, as seen in Figure~\ref{fig:cinn_advection}, and is trainable by gradient descent through the data-fitting loss, as the other neural network weights, while the weights multiplying the spatial components are fixed at $1$. For comparison purposes, we also employ the PINN approach to inverse problems, as outlined in \cite{raissi2019physics}, in which the unknown velocity does not appear in the neural network architecture, but is a separate variable identified by gradient descent optimization of the residue loss function.

  
In this experiment, it is assumed that the initial and boundary conditions are not known, and the data consist only of simulated experimental measurements, which are value of the solution at randomly distributed in the spatial-temporal domain, using Latin hypercube sampling, to which independent zero-mean Gaussian noise of standard deviation $\sigma$ is added. Two noise intensities are considered: $\sigma = 0.01$ (low noise) and $\sigma = 0.05$ (high noise). Both PINN and CINN have the same architecture as in the forward experiment, and Glorot normal initialization with ADAM optimization with a constant learning rate of $0.001$ are used, as previously. The value of the true velocity is set to $v = 1.0$. The (negative of the) velocity weight in the characteristic layer of the CINN and the velocity parameter in the PINN are initialized uniformly in the interval $[0,2]$. As before, the spatial weight in the CINN is non-trainable, and fixed at $1.0$. The PINN uses 300 collocation points, which are randomly distributed in the spatial-temporal domain, using Latin hypercube sampling. 

\begin{figure}[t!]
\centering
\begin{tabular}{cc}
\hspace{2.5em}CINN, low noise  & \hspace{3.5em}CINN, high noise \\
\includegraphics[scale=0.35]{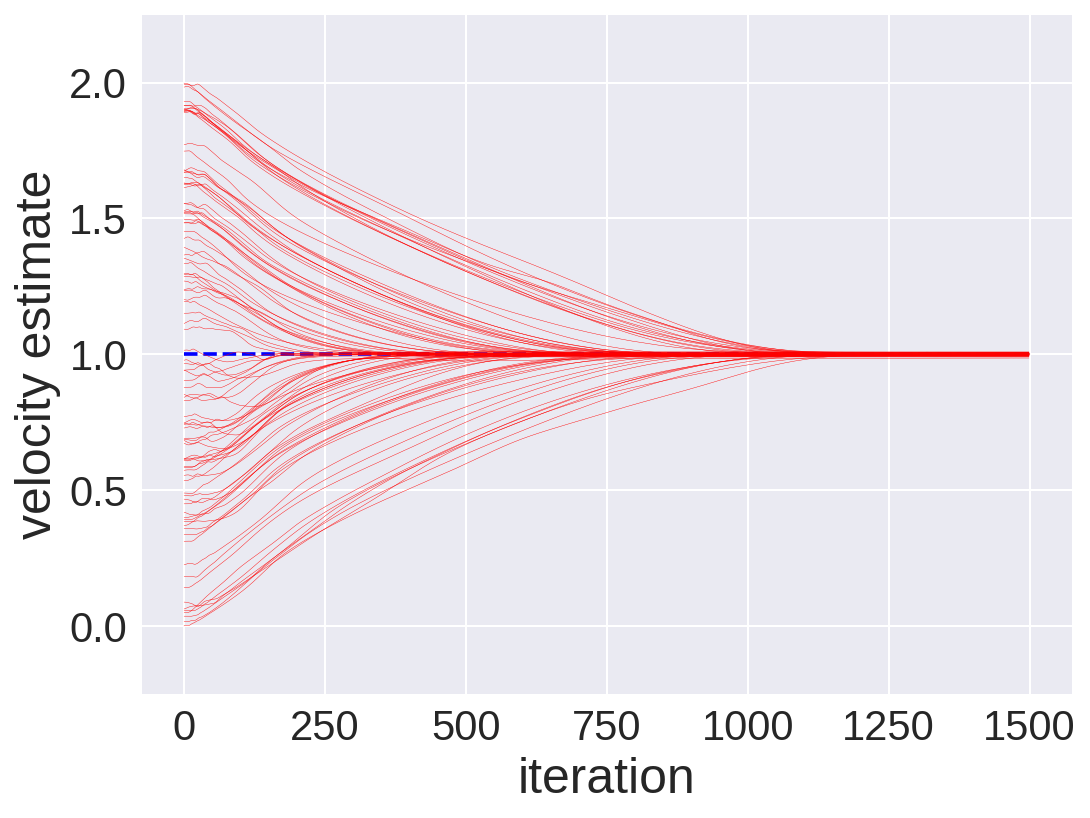} & \hspace{1em}
\includegraphics[scale=0.35]
{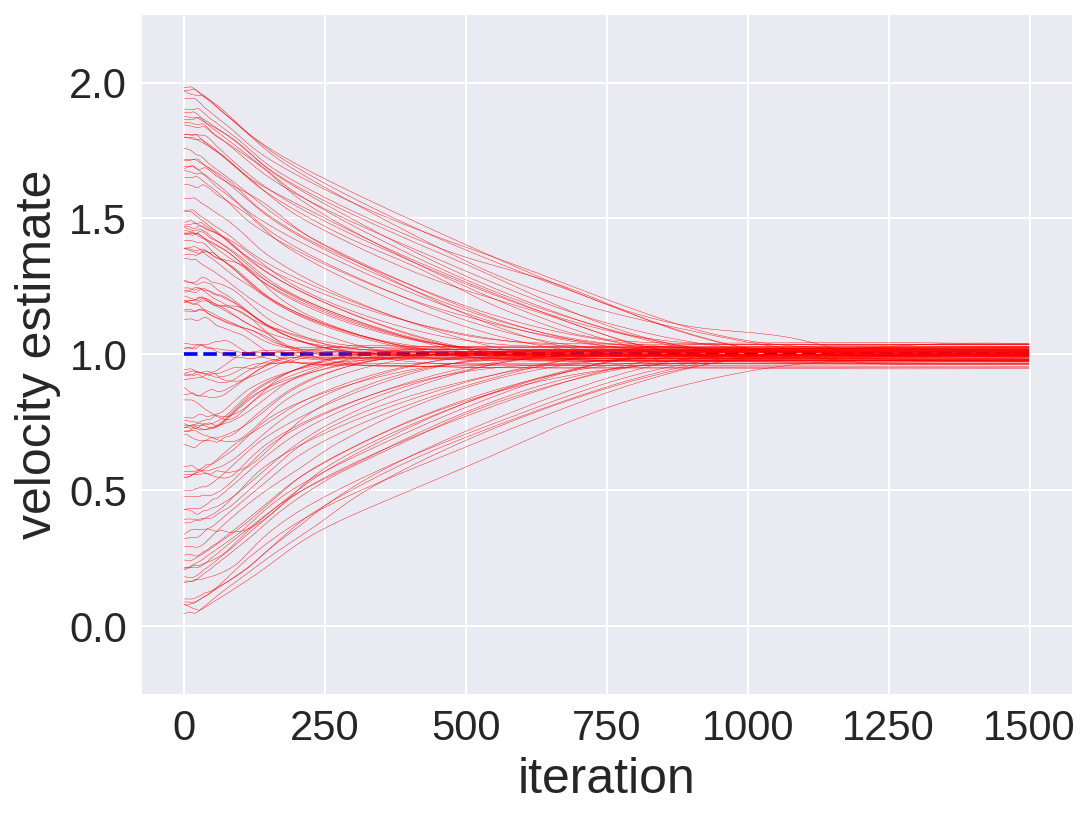} \\[2ex]
\hspace{2.5em}PINN, low noise & \hspace{3.5em}PINN, high noise \\
\includegraphics[scale=0.35]
{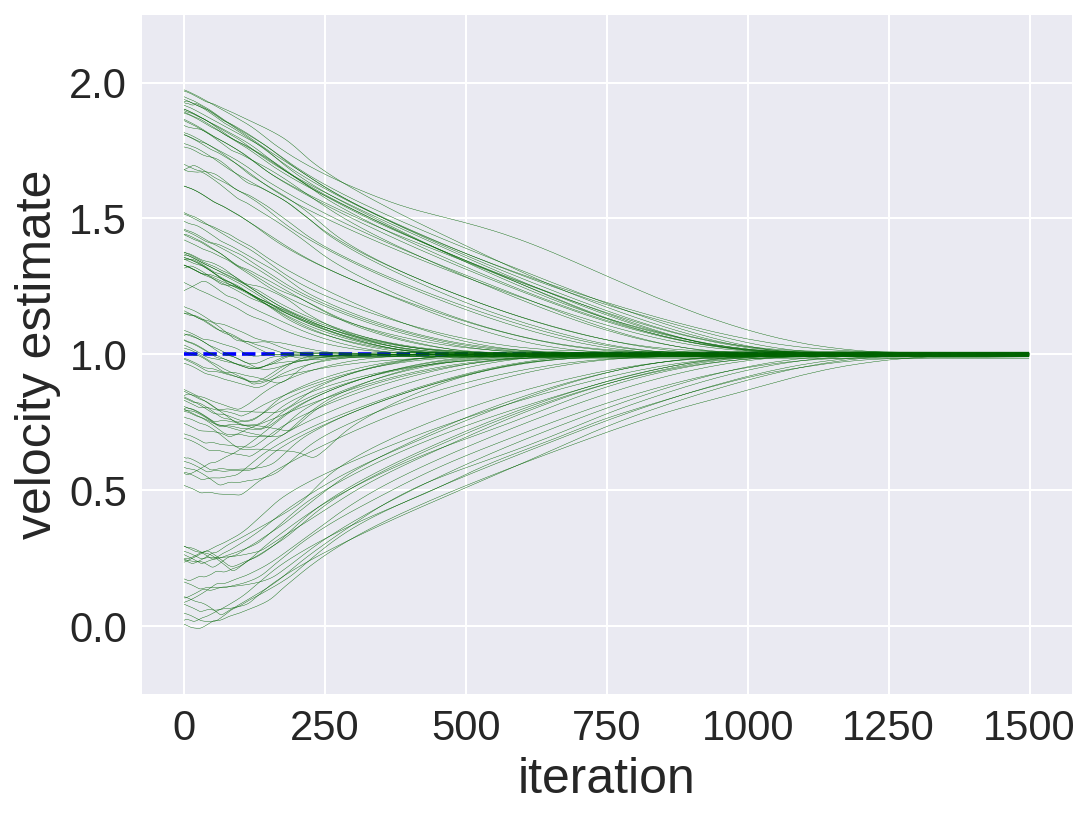} & \hspace{1em}
\includegraphics[scale=0.35]
{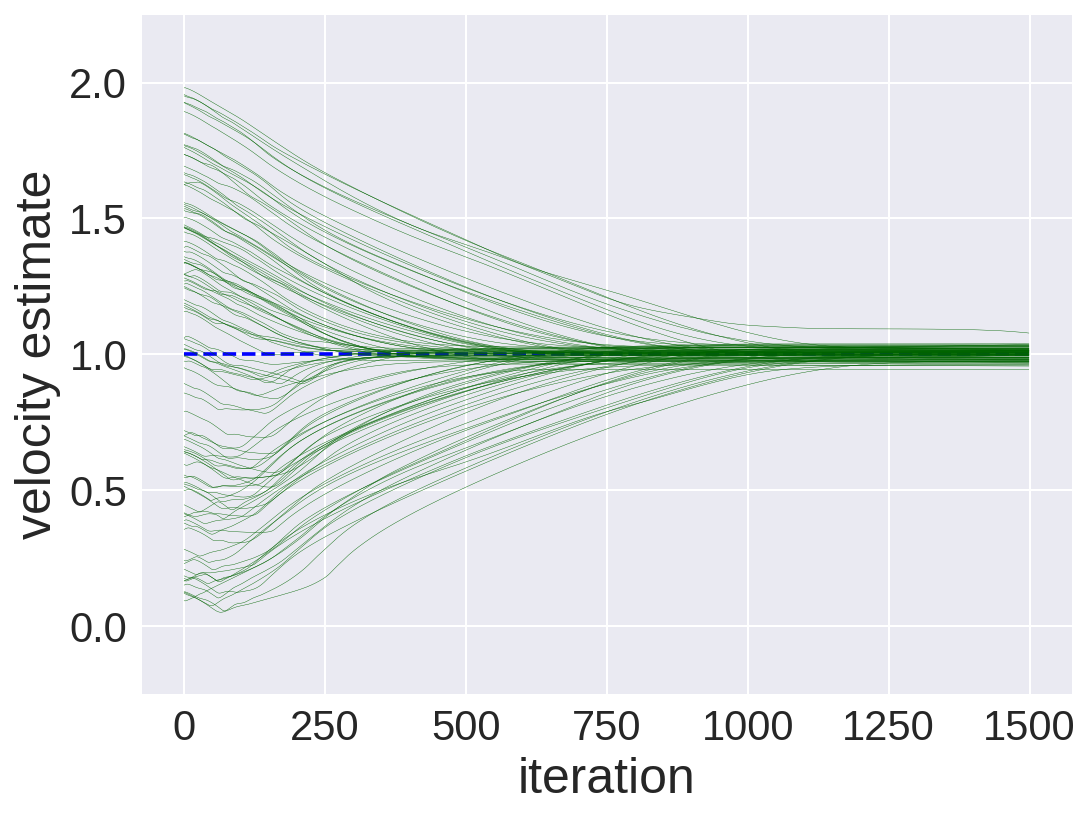}
\end{tabular}
\label{fig:inverse_PINN_CINN_results}
\caption{Velocity estimate evolution in the inverse problem over 100 independent initializations.}
\end{figure}

Figure~\ref{fig:inverse_PINN_CINN_results} displays 
the trajectory of the velocity estimate as a function of iteration number, over 100 independent runs with different random initialization of the neural network weights and velocity estimates. We can see that both the CINN and PINN estimators do a good job. The CINN estimator converges slightly faster and more directly to the correct velocity than the PINN estimator. Quantitative results at the end of the training process are displayed in Table~\ref{tab:advection_inverse}. We can observe that all estimators are nearly unbiased. As indicated by the standard deviation, the CINN estimator is slightly more accurate than the PINN one, in both low and high noise cases. On the other hand, the CINN estimator is significantly faster, as expected.

\begin{table}[h!]
\centering
\begin{tabular}{c|cc}
\hline
Solver & velocity estimate & running time (sec)\\
\hline
CINN, low noise & $1.00030 \pm 0.00366$ & $8.06378 \pm 0.18801$\\[1ex]
CINN, high noise & $0.99967 \pm 0.02045$ & $8.01660 \pm 0.20058$\\[1ex]
PINN, low noise & $0.99949 \pm 0.00421$ & $13.72794 \pm 0.90249$\\[1ex]
PINN, high noise & $1.00170 \pm 0.02117$ & $13.17059 \pm 0.80538$\\[1ex]
\hline
\end{tabular}
\label{tab:advection_inverse}
\caption{Advection inverse problem results over 100 independent repetitions.}
\end{table}

\subsubsection{Advection with Periodic Boundary Conditions}
\label{Sec-periodic}

The experiment in this section shows that CINNs can be trained with periodic boundary conditions using a soft penalty term. The output of this CINN still must satisfy the PDE. The problem under consideration is
\begin{align}
  & \frac{\partial u(x,t)}{\partial t} \:+\: v\, \frac{\partial u(x,t)}{\partial x} \,=\, 0\,, \quad (x,t) \in [0,2\pi] \times [0,T]\,,\\
  & u(x,0) \,=\, \sin(x)\,,\\
  & u(0,t) \,=\, u(2\pi,t)\,, \quad t \in [0,T]\,.\label{eq:initial2}
\end{align}
This problem becomes ``stiffer'' as the advection velocity increases. It was shown in ~\cite{krishnapriyan2021characterizing} that the baseline PINN fail to converge in the case of high stiffness, i.e., high advection velocities. Here we compare the results obtained with a CINN to the results obtained with a baseline PINN using exactly the same hyperparameters. In our experiments, we did not observe a failure to converge, but rather, slower convergence at higher velocities due to stiffness. We observed that the CINN is affected by stiffness, but to a smaller degree than the baseline PINN.

\begin{figure}[t!]
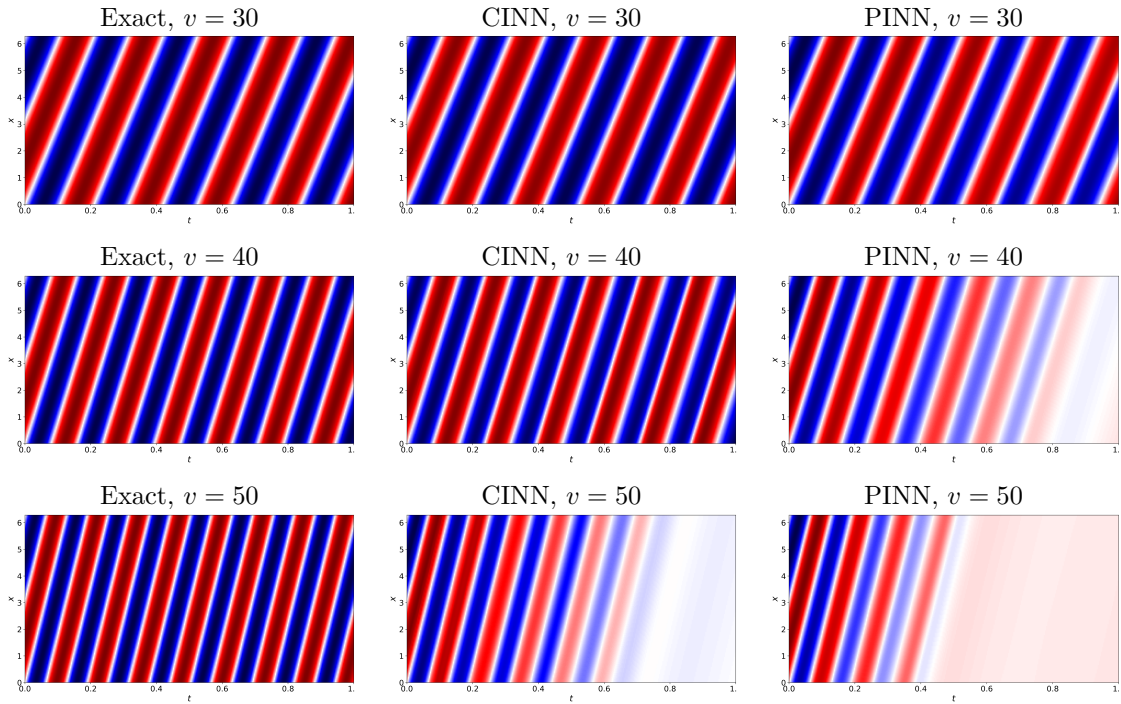

\centering
\begin{tabular}{ccc}
Exact, $v=30$ & CINN, $v=30$ & PINN, $v=30$\\
\adjincludegraphics[scale=0.16,trim={0 0 {.105\width} {0.05\height}},clip]{figs/Exact_Periodic_v30.png} &
\adjincludegraphics[scale=0.16,trim={0 0 {.105\width} {0.05\height}},clip]{figs/CINN_Periodic_v30.png} &
\adjincludegraphics[scale=0.16,trim={0 0 {.105\width} {0.05\height}},clip]{figs/PINN_Periodic_v30.png}\\[0.5ex]
Exact, $v=40$ & CINN, $v=40$ & PINN, $v=40$\\
\adjincludegraphics[scale=0.16,trim={0 0 {.105\width} {0.05\height}},clip]{figs/Exact_Periodic_v40.png} &
\adjincludegraphics[scale=0.16,trim={0 0 {.105\width} {0.05\height}},clip]{figs/CINN_Periodic_v40.png} &
\adjincludegraphics[scale=0.16,trim={0 0 {.105\width} {0.05\height}},clip]{figs/PINN_Periodic_v40.png}\\[0.5ex]
Exact, $v=50$ & CINN, $v=50$ & PINN, $v=50$\\
\adjincludegraphics[scale=0.16,trim={0 0 {.105\width} {0.05\height}},clip]{figs/Exact_Periodic_v50.png} &
\adjincludegraphics[scale=0.16,trim={0 0 {.105\width} {0.05\height}},clip]{figs/CINN_Periodic_v50.png} &
\adjincludegraphics[scale=0.16,trim={0 0 {.105\width} {0.05\height}},clip]{figs/PINN_Periodic_v50.png}
\end{tabular}
\label{fig:periodic_advection_results}
\caption{Periodic advection results after 20,000 ADAM iterations. The worst result over 10 repetitions, according to $L_2$ error, is displayed for the CINN and PINN.}
\end{figure}

\begin{table}[h!]
\centering
\begin{tabular}{c|c|ccc}
\hline
Solver & $\quad v \quad$ & $L_1$ error & $L_2$ error & running time (sec)\\
\hline
\multirow{3}{*}{CINN} & $20$ & $0.0253 \pm 0.0036$ & $0.0300 \pm 0.0043$ & $97.1958 \pm 1.1487$\\[1ex]
& 30 & $0.0478 \pm 0.0072$ & $0.0579 \pm 0.0095$ & $97.6086 \pm 1.6725$\\[1ex]
& 40 & $0.0714 \pm 0.0148$ & $0.0852 \pm 0.0169$ & $100.8676 \pm 1.1128$\\[1ex]
& 50 & $0.4692 \pm 0.1533$ & $0.5365 \pm 0.1729$ & $100.2374 \pm 0.8882$\\[1ex]
\hline
\multirow{3}{*}{PINN} & 20 & $0.0305 \pm 0.0062$ & $0.0347 \pm 0.0056$ & $119.1307 \pm 1.0829$\\[1ex]
& 30 & $0.0873 \pm 0.0162$ & $0.1003 \pm 0.0188$ & $117.8713 \pm 1.7974$\\[1ex]
& 40 & $0.3918 \pm 0.1002$ & $0.4395 \pm 0.1120$ & $117.9513 \pm 1.7525$\\[1ex]
& 50 & $0.7140 \pm 0.0278$ & $0.7797 \pm 0.0242$ & $129.9164 \pm 2.1761$\\[1ex]
\hline
\end{tabular}
\label{tab:periodic_advection_forward}
\caption{Periodic advection forward prediction results over 10 independent repetitions.}
\end{table}

Our experiment employs $200$ uniformly distributed data points along the initial boundary $t=0$. The periodic boundary condition is enforced softly by means of the loss:
\begin{equation}
    \mathcal{L}_b(\theta) \,=\, \frac{1}{n_b}\sum_{i=1}^{n_b} |u_\theta(0,t_b^i)-u_\theta(2\pi,t_b^i)|^2,
\label{eq-Lb}
\end{equation}
where $t_b^1,\ldots,t_b^{n_b}$ are time points in the interval $[0,T]$. In our experiment, $n_b=1000$ uniformly distributed time points are used. The baseline PINN employs the same initial and boundary points as the CINN, plus a large set of $10,000$ collocation points randomly distributed in the spatial-temporal domain using Latin hypercube sampling. Both the CINN and the PINN consist of 4 layers of 50 neurons each, with hyperbolic tangent nonlinearity, which is the architecture used in~\cite{krishnapriyan2021characterizing} (the CINN contains in addition the initial characteristic layer). Glorot initialization for the trainable weights was employed and 20,000 iterations of ADAM gradient descent, with the usual parameters, were performed. A total of 10 independent replication of the experiment were conducted, to account for the random initialization of the neural network weights. Figure~\ref{fig:periodic_advection_results} displays 
worst-case results (according to $L_2$ error) obtained in the experiment.  

One can see that, as the advection velocity increases and the PDE becomes stiffer, convergence to the solution becomes slower, but the convergence of the CINN is faster than that of PINN. The quantitative results displayed in Table~\ref{tab:periodic_advection_forward} confirm this. At a slow velocity, the CINN and PINN perform similarly, other than the training speed; this is the same behavior as in the earlier simple advection experiment. As $v$ increases, the $L_1$ and $L_2$ errors become larger, but here it is clear that the PINN errors increase much faster than the CINN errors. The critical case is $v=40$, when, at the end of 20,000 training iterations, the CINN still reaches close to the solution, but the PINN fails to get near it, as seen in the second row of Figure~\ref{fig:periodic_advection_results}, and the CINN produces errors that are more than 5 times smaller than the PINN does, as seen in Table~\ref{tab:periodic_advection_forward}. In all cases, the CINN is still faster to train the PINN, even though the difference is smaller than before, due to the presence of the same extra boundary loss term in both CINN and PINN.

Finally, we observe that in~\cite{krishnapriyan2021characterizing}, only the results for $v=30$ are shown, when the PINN has not converged; but as the results here show, the velocity at which this happens depends on the number of training iterations. In~\cite{krishnapriyan2021characterizing}, a ``curriculum training'' approach is used to obtain good results at $v=30$, whereby $v$ is slowly increased to $30$ and PINN retrained each time. This approach is clearly very time-consuming, and could become impractical in very large problems.


\subsection{Linear System of PDEs}

\begin{figure}[t!]
\centering
\includegraphics[scale=0.6]{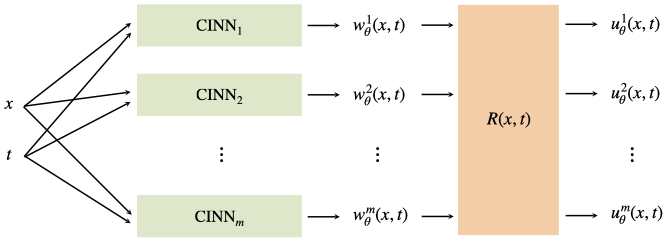}
\caption{CINN architecture for a linear hyperbolic PDE system. This model can be trained end-to-end from sample data using standard deep learning software.}
\label{fig:cinn_system}
\end{figure}

The extension of the linear transport equation \eqref{eq:tr2} to a one-dimensional linear system of PDEs is of the form:
\begin{equation}
  \frac{\partial \bu(x,t)}{\partial t} \:+\: A(x,t)\,\, \frac{\partial \bu(x,t)}{\partial x} \,=\, 0\,,
\label{eq:system}
\end{equation}
where $\bu:R \rt R^m$ and $A(x,t)$ is an $m \times m$ matrix. This represents a system of $m$ PDEs in $m$ variables $\bu = (u^1,\ldots,u^m)$. This problem is {\em hyperbolic}, i.e., it produces wave-like solutions, if the matrix $A(x,t)$ is diagonalizable with real eigenvalues at all points $(x,t)$ in the domain~\cite{leveque2002finite}. In this case, one can write
\beq
  A(x,t) \,=\, R(x,t)\, \Lambda(x,t)\, R(x,t)^{-1},
\eeq  
where $\Lambda(x,t)$ is the diagonal matrix of eigenvalues $\lambda_1(x,t),\ldots,\lambda_m(x,t)$. Introducing the characteristic variables $\bw = (w^1,\ldots,w^m)$ by means of the transformation
\beq
  \bw(x,t) \,=\, R(x,t)^{-1}\bu(x,t)
\eeq
allows one to rewrite \eqref{eq:system} as
\begin{equation}
  \frac{\partial \bw(x,t)}{\partial t} \:+\: \Lambda(x,t)\, \frac{\partial \bw(x,t)}{\partial x} \,=\, 0\,.
\label{eq:system2}
\end{equation}
This is a system of $m$ decoupled advection equations:
\begin{equation}
  \frac{\partial w^i(x,t)}{\partial t} \:+\: \lambda_i(x,t)\, \frac{\partial w^i(x,t)}{\partial x} \,=\, 0\,, \quad i=1,.\ldots,m\,.
\label{eq:system3}
\end{equation}
Note that these equations are not simple linear advections; in particular, their characteristic curves are not necessarily lines. however, each of these equations can be solved by a separate CINN, as described in Section~\ref{Sec-linscalar}. The outputs of these CINNs are multiplied by the matrix $R(x,t)$ to return to the original variables. This architecture is summarized in Figure~\ref{fig:cinn_system}. Notice that this model can be trained end-to-end using data on the original variable $\bu$. In addition, it can be implemented easily using the functional API of Keras in Tensorflow 2.x.

\subsubsection{Hidden Acoustics Problem}

We demonstrate the use of the previous CINN architecture in a ``hidden'' acoustics experiment, where the classical equations of linear acoustics and scattered data on one of the fields (in this case, the velocity), are used to predict the other, unseen field (in this case, the pressure). It will be seen below that CINN produced significantly more accurate results than a comparable baseline PINN.

The equations of linear acoustics in a gas in a one-dimensional tube are:
\begin{align}
   &\frac{\partial p(x,t)}{\partial t}\:+\: K_0\, \frac{\partial v(x,t)}{\partial x}  \,=\, 0\,,\\
  &\frac{\partial v(x,t)}{\partial t}\:+\: \frac{c_0^2}{K_0} \, \frac{\partial p(x,t)}{\partial x}  \,=\, 0\,,
  \end{align}
where  $p(x,t)$ and $v(x,t)$ are small pressure and velocity disturbances (i.e., sound) around a fixed pressure and velocity background state, respectively (here, the background velocity is assumed, without loss of generality, to be zero), while the constants $K_0$ and $c_0$ are the {\em bulk modulus of compressibility} of the gas and the speed of sound in the gas, respectively. The derivation of these equations from first principles of conservation of mass and momentum can be found in~\cite{leveque2002finite}.

The equations of linear acoustics can be rewritten as 
\begin{equation}
  \frac{\partial}{\partial t}\bbm p(x,t) \\ v(x,t) \ebm \:+\: \bbm 0 & \!\! K_0\\ c_0^2/K_0 & \!\! 0 \ebm \frac{\partial}{\partial x} \bbm p(x,t) \\ v(x,t) \ebm  =\, 0\,,
\label{eq:system_acc}
\end{equation}
which is in the form~\eqref{eq:system}, with 
\beq
  \bu(x,t) \,=\, \bbm p(x,t) \\ v(x,t) \ebm, \quad A \,=\, \bbm 0 & \!\! K_0\\ c_0^2/K_0 & \!\! 0 \ebm\,.
\eeq
This is therefore a constant-coefficient system, with $A(x,t) \equiv A$. It is easy to see that
\beq
  \Lambda \,=\, \bbm -c_0 & 0 \\ 0 & c_0 \ebm, \quad R \,=\, \bbm -Z_0 & \!\! Z_0\\ 1 & \!\!1 \ebm\,,
\eeq
where $Z_0 = K_0/c_0$ is the {\em impedance} of the medium. The uncoupled equations in the characteristic variables are simple advections:
\begin{equation}
  \frac{\partial w^i(x,t)}{\partial t} \:+\: (-c_0)^i\, \frac{\partial w^i(x,t)}{\partial x} \,=\, 0\,, \quad i=1,2\,.
\label{eq:system2_acc}
\end{equation}
As seen previously, the solutions are of the form
\beq
   w^1(x,t) \,=\, g(x+c_0t) \,, \quad\quad 
   w^2(x,t) \,=\, f(x-c_0t) \,, 
 \eeq  
 where $f$ and $g$ are arbitrary functions. The general solution in the original variables is $\bu(x,t) =R \bw(x,t)$, which gives
 \beq
 \begin{aligned}
   &p(x,t) \,=\, Z_0(f(x-c_0t) - g(x+c_0t))\\
   &v(x,t) \,=\, f(x-c_0t) + g(x+c_0t)\,,
 \end{aligned}
 \label{eq-pv}
 \eeq
 i.e., both pressure and velocity are the superposition of waves moving to the left and right with the sound of speed $c_0$. The wave shapes $f$ and $g$ are determined by initial and/or boundary conditions on the pressure and velocity (or, in the case of a CINN, they are learned from data).

Here, we consider a very long tube, so that boundary conditions do not come into play, and let the initial conditions on the two fields be
 \beq
 \begin{aligned}
 &p(x,0) \,=\, e^{-100x^2},\\
  & v(x,0) \,=\, 0\,,
 \end{aligned}
 \eeq
that is, at first the gas is perfectly still, and then there is a sudden narrow ``bump'' in pressure around $x=0$.
 Letting $t=0$ in \eqref{eq-pv} leads to:
 \begin{align}
  & f(x) \,=\, \frac{1}{2Z_0} e^{-100x^2}\\
   & g(x) \,=\, -f(x) \,=\, -\frac{1}{2Z_0} e^{-100x^2}\,,
 \end{align}
and, finally, the solution is
\beq
\begin{aligned}
   &p(x,t) \,=\, \frac{1}{2}\left[e^{-100(x-c_0t)^2} + e^{-100(x+c_0t)^2}\right],\\
   &v(x,t) \,=\,  \frac{1}{2Z_0}\left[e^{-100(x-c_0t)^2} - e^{-100(x+c_0t)^2}\right].
\end{aligned}
\label{eq-pvsol}
\eeq
In the experiment, we assume that scattered data only on the velocity field is available, and seek to predict the unseen pressure field using a CINN, which is again compared to a baseline PINN for the same problem. The velocity data is assumed to be distributed randomly in the spatial-temporal domain using Latin hypercube sampling; in particular, no data on the initial conditions are used. A total of 200 points sampled from the true velocity field were used. The values used for the constants are $c_0 = Z_0 = 1$. The CINN follows the architecture in Figure~\ref{fig:cinn_system}, with eight layers of 20 neurons in each of the $m=2$ component branches, in addition to the characteristic layers and the output layer, with hyperbolic tangent nonlinearity, and Glorot initialization for the trainable weights. The network is trained with the standard MSE data-fitting loss function, using ADAM gradient descent with the usual parameters, and a constant learning rate of $0.001$. A comparable PINN is used, consisting of 8 layers of 40 neurons each (to keep the number of neurons about the same in both cases), and 300 collocation points randomly distributed in the spatio-temporal domain using Latin hypercube sampling.

\begin{figure}[t!]
\centering
\begin{tabular}{ccc}
CINN after 1000 training iterations\\
\includegraphics[scale=0.48]{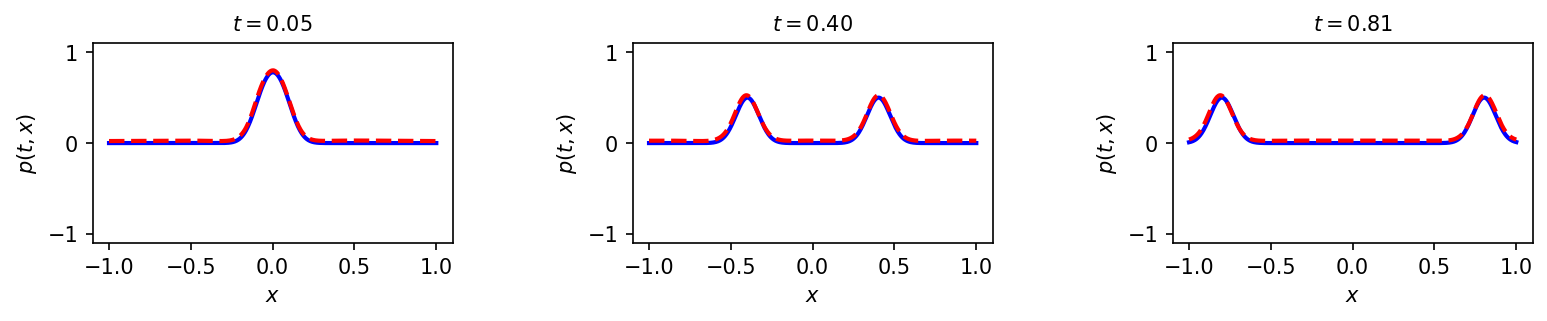}\\
\includegraphics[scale=0.48]{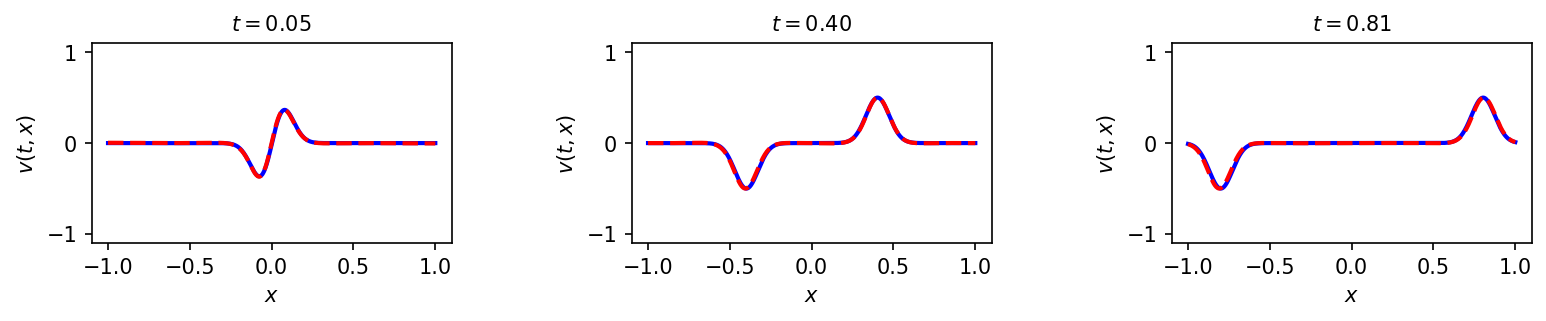}\\[1ex]
PINN after 1000 training iterations\\
\includegraphics[scale=0.48]{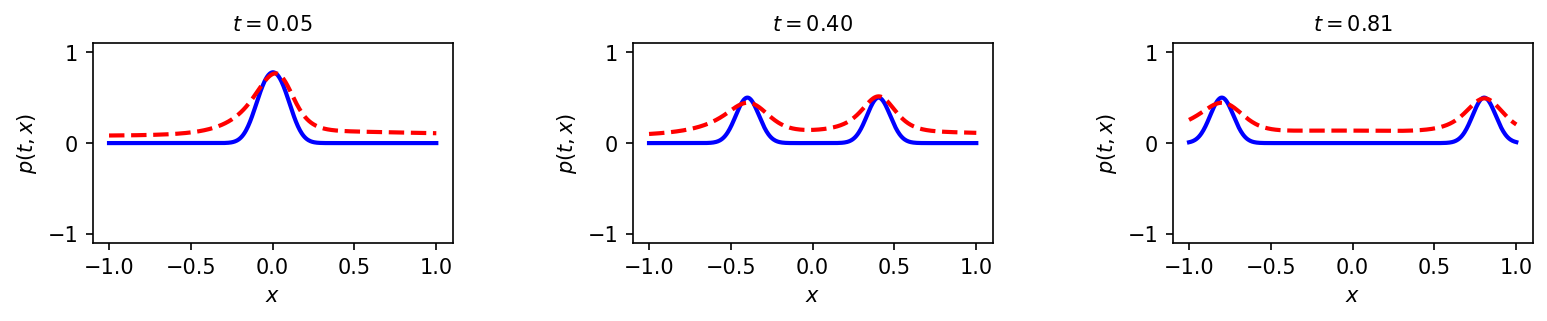}\\
\includegraphics[scale=0.48]{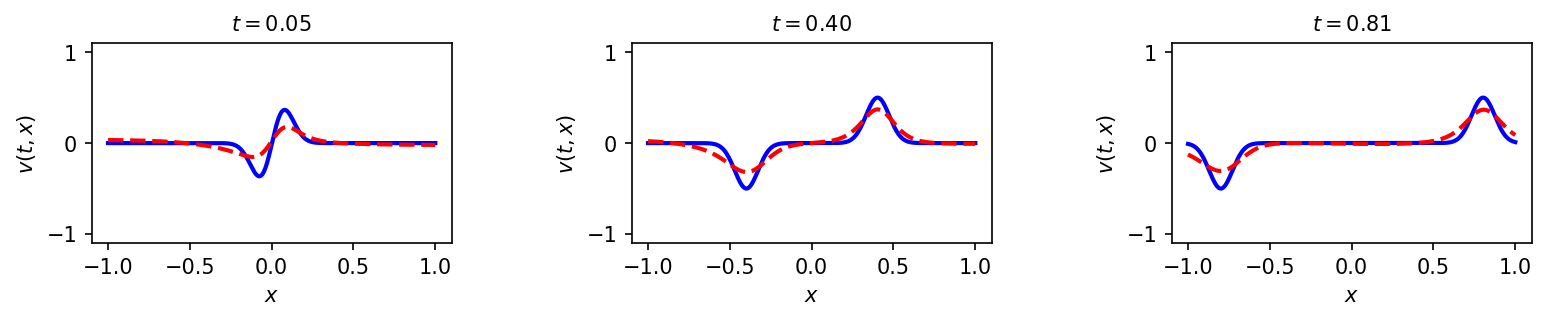}\\[1ex]
PINN after 5000 training iterations\\
\includegraphics[scale=0.48]{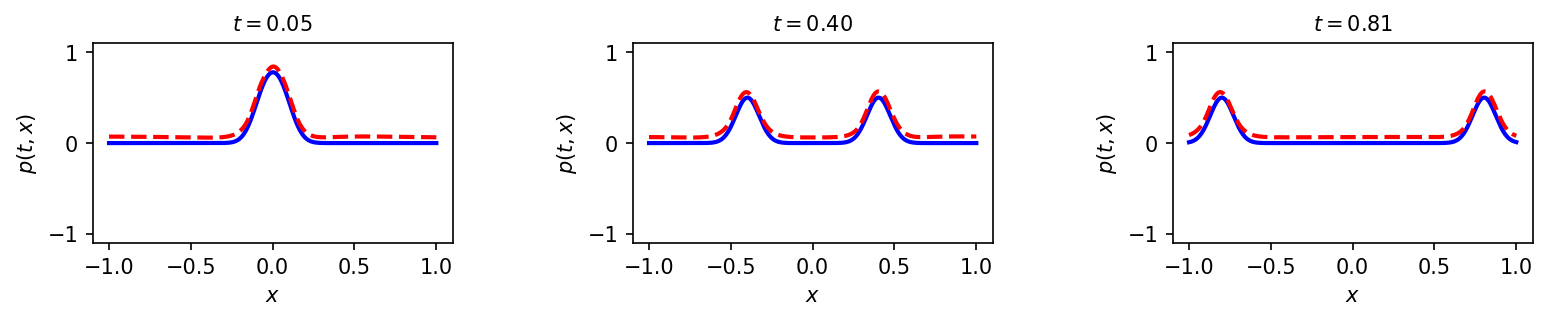}\\
\includegraphics[scale=0.48]{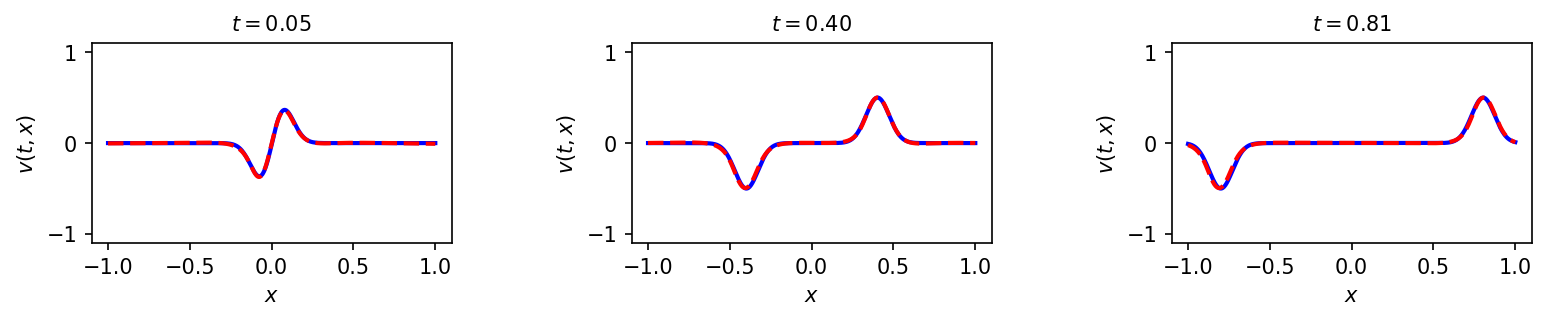}\\[1ex]
\end{tabular}
\label{fig:acoustics_results}
\caption{Typical results from the hidden acoustics experiment.}
\end{figure}

Typical results can be seen in Figure~\ref{fig:acoustics_results}. Convergence of the CINN prediction of both the pressure and velocity fields after 1000 training iterations is quite good. We reiterate that, at any time during training, the CINN velocity and pressure outputs are a solution of the acoustics equations. For the comparable baseline PINN, we can observe that 1000 iterations is not enough to obtain good results. At 5000 iterations, the prediction of the velocity field is good, but the prediction of the unseen pressure field is still noticeably off.

The quantitative results in Table~\ref{tab:hidden_acoustics} confirm the previous observations (only the $L_2$ results are shown to make the table concise; the $L_1$ errors show the same trends). In all cases, prediction of the velocity field are more accurate than those for the unseen pressure field, as expected. However, the CINN is significantly more accurate than the PINN in predicting the pressure field, in all cases. The CINN is also more accurate in predicting the velocity field, unless 5000 training iterations are used with the PINN, at a significant computational cost.


\section{Extension to NonLinear PDEs}
\label{Sec-nonlin}

Experimental results with linear, divergence-free scalar transport equation, in both the scalar and system cases, showed that the CINN approach is fast and effective. In this section, we discuss briefly below an extension of the CINN approach to the nonlinear transport PDEs.

Consider the nonlinear conservation law:
\beq
   \frac{\partial u(x,t)}{\partial t} \:+\: \frac{\partial F(u(x,t))}{\partial x} \,=\, 0\,,
\label{eq:conserv2}
\eeq
where $F:R \rt R$ is the flux. Using the chain rule, this can be rewritten as:
\beq
   \frac{\partial u(x,t)}{\partial t} \:+\: F'(u(x,t))\, \frac{\partial u(x,t)}{\partial x} \,=\, 0\,.
\label{eq:conserv3}
\eeq
Comparing this to \eqref{eq:tr2}, we observe that this can be seen as a (one-dimensional) transport equation where the velocity field $v(u) = F'(u)$ depends on the the unknown $u$ itself (though it does not depend on $x$ and $t$ directly). A classic example is the inviscid Burgers equation, where $F(u) = \frac{1}{2}u^2$, so that $v(u) = F'(u) = u$.

\begin{table}[t!]
\centering
\begin{tabular}{c|c|ccc}
\hline
Solver & No. iterations & pressure $L_2$ error & velocity $L_2$ error & running time (sec)\\[1ex]
\hline
CINN & $1000$ & $0.1267 \pm 0.0534$ & $0.0588 \pm 0.0077$ & $9.0734 \pm 0.6197$\\[1ex]
\hline
\multirow{2}{*}{PINN}
& $1000$ & $0.7174 \pm 0.1994$ & $0.5392 \pm 0.1866$ & $13.1619 \pm 0.7929$\\[1ex]
& $3000$ & $0.7017 \pm 0.4271$ & $0.2287 \pm 0.2451$ & $29.1454 \pm 1.1361$\\[1ex]
& $5000$ & $0.5209 \pm 0.2774$ & $0.0615 \pm 0.0075$ & $44.4225 \pm 0.6568$\\[1ex]
\hline
\end{tabular}
\label{tab:hidden_acoustics}
\caption{Hidden acoustics experiment prediction results over 10 independent repetitions.}
\end{table}

\begin{figure}[t!]
\centering
\includegraphics[scale=0.67]{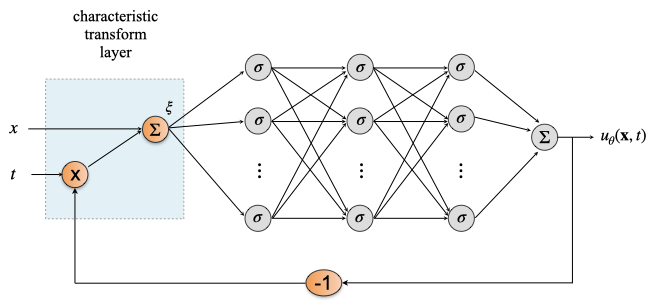}
\caption{Proposed recursive CINN architecture to solve a scalar nonlinear conservation law.}
\label{fig:cinn_nonlinear}
\end{figure}

Consider a characteristic curve $x(t)$; $t \geq 0$ defined as the solution of
the ODE:
\beq
  \frac{d x(t)}{dt} \,=\, v(u(x(t),t))\,.
\label{eq:charac3}
\eeq
Along this curve, we have
\beq
\bal
  \frac{du(x(t),t)}{dt} & \,=\,  \frac{\partial u(x(t),t)}{\partial t} + \frac{d x(t)}{dt} \, \frac{\partial u(x,t)}{\partial x} \\
  & \,=\, \frac{\partial u(x(t),t)}{\partial t} + v(u(x(t),t)) \, \frac{\partial u(x,t)}{\partial x} \\
  & \,=\, 0\,.
\eal
\eeq
Therefore, $u(x(t),t)$ has a constant value $\eta$ along $x(t)$, and \eqref{eq:charac3} becomes
\beq
  \frac{d x(t)}{dt} \,=\, \eta\,,
\label{eq:charac4}
\eeq
with general solution
\beq
  x(t) \,=\, \eta t + \xi\,,
\eeq  
i.e., the characteristic curve $x(t)$ is a line with slope $\eta$, which is determined by the data in the problem. Unlike the linear case, these characteristics are not parallel to each other, since the slope  $\eta$ is in general different for different characteristics. This leads to {\em compression} and {\em rarefaction} waves, when characteristics converge and diverge from each other, respectively, a phenomenon that is non-existent in the linear case. Characteristics may indeed cross, when the solution would become multi-valued. Hence, there can be no solution at that point, in the classic sense. However, a solution with a {\em shock} is well-defined, in a weak sense. (See \cite{leveque2002finite} for more details.)  

As in the linear case, one can define a characteristic transform
\beq
  \phi(\bx,t,\eta) \,=\, x(t) - \eta t \,=\, \xi\,.
\label{eq:chtf2}
\eeq 
By analogy with the CINN architecture for the linear advection equation in Figure~\ref{fig:cinn_advection}, we propose the recursive architecture in Figure~\ref{fig:cinn_nonlinear} to solve this nonlinear PDE. As with every recursive neural network, this architecture can be trained to convergence by ``unrolling". More complex nonlinear hyperbolic PDEs could be solved in similar fashion.

\section{Conclusion}

A standard neural network is agnostic to the physics of the problem, and therefore requires a large amount of data to make reliable predictions. On the other hand, the PINN approach includes the physics in the training of the neural network by adding a PDE residual loss term based on a set of collocation points. However, this still has two main drawbacks. First, the PDE is never guaranteed to be satisfied over the entire domain, even if the residual term is zero, since it is only enforced at a finite set of collocation points; in practice, the residual is nonzero, and the PDE is not satisfied even at the collocation points. This leads to the occurrence of non-physical solutions, e.g., conservation laws may be  broken. The second drawback is less obvious: the PINN loss function is a scalarization into a single objective of a multiple-objective optimization problem. In reality, there are two objectives, fitting the data and satisfying the PDE, rendering PINN training difficult. In this preprint, we have proposed CINN, a simple approach that does away with all these issues: the PDE is satisfied without the need of a collocation approach, and training is based on a standard single-objective data-fitting loss function. The main advantages of CINNs are its simplicity and efficiency, in addition to the avoidance of non-physical solutions. Indeed, the preliminary empirical results presented here demonstrate that CINNs can be significantly faster to train and more accurate than comparable baseline PINNs. It was seen that this efficiency advantage increases as the complexity of the problem increased, from the simple advection, to a periodic advection, to the hidden acoustics problem. This trend is expected to continue in high-dimensional, complex problems. Future work will investigate such problems, in addition to extensions to variable-coefficient and nonlinear PDEs, presence of diffusion, and more.

\section*{Acknowledgement}
This work was supported by the National Science Foundation under award CCF-2225507. 

\bibliographystyle{ieeetr}
\bibliography{ArXiV_v2}

\end{document}